\def\year{2023}\relax
\documentclass[letterpaper]{article} % DO NOT CHANGE THIS
\usepackage{aaai22}  % DO NOT CHANGE THIS
\usepackage{times}  % DO NOT CHANGE THIS
\usepackage{helvet}  % DO NOT CHANGE THIS
\usepackage{courier}  % DO NOT CHANGE THIS
\usepackage[hyphens]{url}  % DO NOT CHANGE THIS
\usepackage{graphicx} % DO NOT CHANGE THIS
\usepackage{natbib}  % DO NOT CHANGE THIS AND DO NOT ADD ANY OPTIONS TO IT
\usepackage{caption} % DO NOT CHANGE THIS AND DO NOT ADD ANY OPTIONS TO IT
\DeclareCaptionStyle{ruled}{labelfont=normalfont,labelsep=colon,strut=off} % DO NOT CHANGE THIS
\usepackage{algorithm}
\usepackage{algorithmic}

\usepackage{xspace}
\usepackage{booktabs}
\usepackage{amsmath}
\usepackage{amssymb}

\usepackage{newfloat}
\usepackage{listings}
\floatstyle{ruled}
\newfloat{listing}{tb}{lst}{}
\floatname{listing}{Listing}
\title{Enhancing Large Language Models with Climate Resources}
\author{
    Mathias Kraus,\textsuperscript{\rm 1}
    Julia Anna Bingler,\textsuperscript{\rm 2,5}
    Markus Leippold,\textsuperscript{\rm 3}
    Tobias Schimanski,\textsuperscript{\rm 3}
    Chiara Colesanti Senni,\textsuperscript{\rm 3,5}
    Dominik Stammbach,\textsuperscript{\rm 4}
    Saeid Ashraf Vaghefi,\textsuperscript{\rm 3}
    Nicolas Webersinke\textsuperscript{\rm 1}
}
\affiliations{
    \textsuperscript{\rm 1}FAU Erlangen-Nuremberg, Germany\\
    \textsuperscript{\rm 2}University of Oxford, United Kingdom\\
    \textsuperscript{\rm 3}University of Zurich, Switzerland\\
    \textsuperscript{\rm 4}ETH Zurich, Switzerland\\
    \textsuperscript{\rm 5}Council on Economic Policies\\
    mathias.kraus@fau.de, julia.bingler@smithschool.ox.ac.uk, markus.leippold@bf.uzh.ch, tobias.schimanski@bf.uzh.ch, chiara.colesantisenni@bf.uzh.ch, dominik.stammbach@gess.ethz.ch, saeid.vaghefi@geo.uzh.ch, nicolas.webersinke@fau.de 
}

\begin{document}

\maketitle

\begin{abstract}
Large language models (LLMs) have significantly transformed the landscape of artificial intelligence by demonstrating their ability in generating human-like text across diverse topics. However, despite their impressive capabilities, LLMs lack recent information and often employ imprecise language, which can be detrimental in domains where accuracy is crucial, such as climate change. In this study, we make use of recent ideas to harness the potential of LLMs by viewing them as agents that access multiple sources, including databases containing recent and precise information about organizations, institutions, and companies. We demonstrate the effectiveness of our method through a prototype agent that retrieves emission data from ClimateWatch (https://www.climatewatchdata.org/) and leverages general Google search. By integrating these resources with LLMs, our approach overcomes the limitations associated with imprecise language and delivers more reliable and accurate information in the critical domain of climate change. This work paves the way for future advancements in LLMs and their application in domains where precision is of paramount importance.
\end{abstract}

\section{Introduction}

\begin{figure*}[h!]
	\centering
	\includegraphics[width=0.8\textwidth]{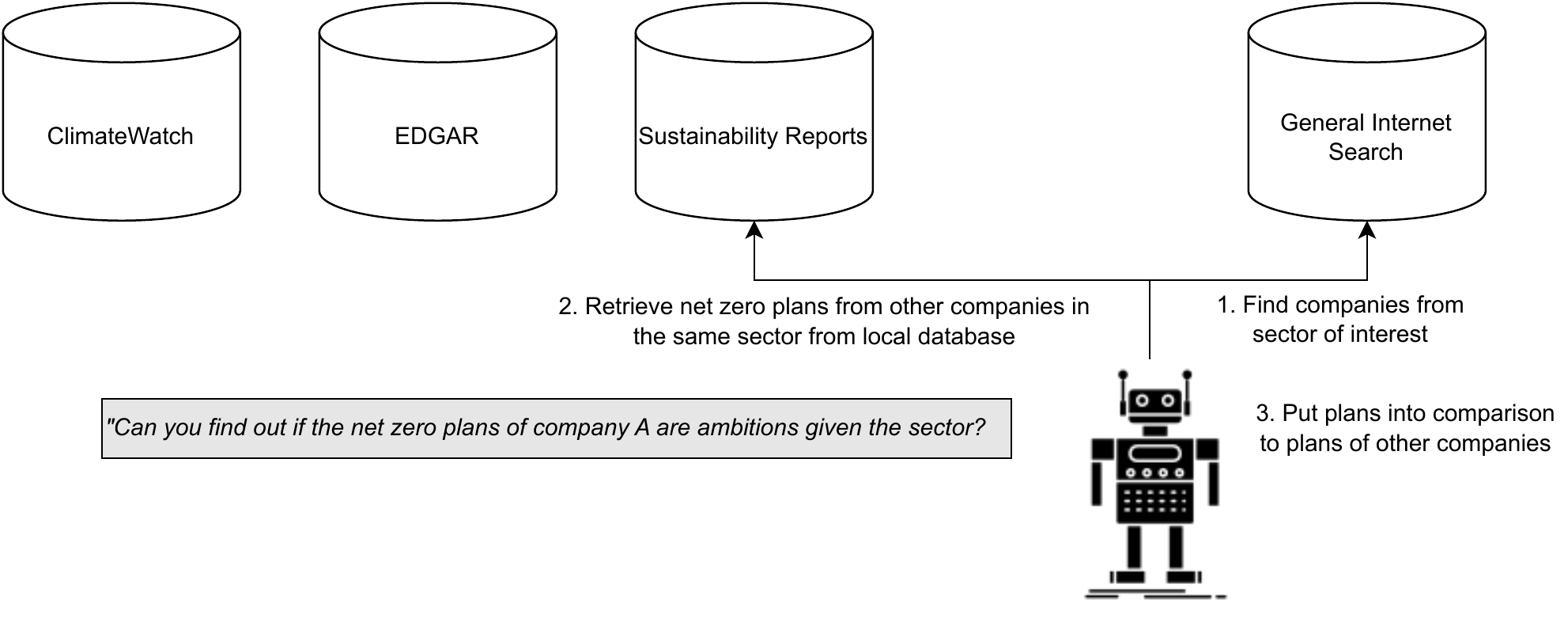}
	\caption{Setup for LLM agents that can access multiple sources, such as databases or general internet search.}
	\label{fig:llm_climate}
\end{figure*}

\textbf{Motivation.} Large language models have revolutionized the field of artificial intelligence (AI) in recent years \cite{brown2020language,radford2019language,ouyang2022training}. Models such as T0 \cite{sanh2021multitask}, LLama \cite{touvron2023llama}, Palm \cite{chowdhery2022palm}, GPT-3 \cite{brown2020language}, GPT-4 \cite{gpt4techreport}, or instruction finetuned models, such as ChatGPT \cite{instructgpt} have demonstrated their exceptional capabilities in generating human-like text across various domains, including language translation, summarization, and question answering. However, despite their impressive performance, LLMs are not without limitations. First, LLMs by themselves only have access to information from their training data (i.e., data from the past). Thus LLMs can not access the most recent information which prohibits usage in the domain of climate change to a large extend. Another critical drawback is their tendency to use imprecise language, which can lead to unreliable or inaccurate predictions in domains where precision and responsibility are crucial. Thus, criticism of these systems has grown, with a recent letter from several experts in the field asking, "Should we let machines flood our information channels with propaganda and untruth?" \cite{FLI2023}. 

One such domain where precision and accuracy are paramount is research on climate reporting \cite{board2017task}. Accurate data and information are essential for policymakers, scientists, and the general public to understand the magnitude and urgency of the climate crisis and take necessary actions to mitigate its impact \cite{tcfd_overview}. Inaccurate information, on the other hand, can lead to inappropriate policies or delayed actions, exacerbating the consequences of climate change \cite{stern2007economics}. However, retrieving and interpreting accurate data from a vast amount of climate-related information is a challenging task \cite{bingler2022cheap}. One example of a valuable source of relevant information for understanding a firm's climate-related performance is annual reports, which contain both quantitative and qualitative information \cite{edmans2016realcosts}. However, this type of information is not naturally represented in LLMs and must be retrieved separately.

\textbf{Contribution.} To address this challenge, we use recent developments in LLMs as agents that access multiple sources, including databases containing precise information about organizations, institutions, and companies. By integrating these resources with LLMs, our approach mitigates the limitations associated with imprecise language, providing more reliable and accurate information in the critical domain of climate change. We present a prototype LLM agent that retrieves emission data from ClimateWatch (\url{https://www.climatewatchdata.org/}) and leverages general Google search. Through two exemplary experiments, we demonstrate the potential of LLMs in the area of climate change. We anticipate that our approach will encourage further research on integrating LLMs with accurate data sources, ultimately fostering more responsible and reliable AI systems.

\textbf{Findings.} Our study demonstrates the potential of LLMs as agents that access multiple sources of information to deliver reliable and accurate information in the domain of climate change. We have shown how our prototype agent can retrieve and process data from ClimateWatch, a database containing precise information on emissions from countries. Additionally, our approach leverages general Google search to complement the information retrieved from ClimateWatch and provide additional context to the generated text. Our experiments and analyses showcase how LLM agents can be used to access multiple datasources and combine this information to provide accurate and reliable responses.

\textbf{Implications.} The implications of our study are significant for both practice and research. In practice, this work is not limited to the field of climate change but can be used to improve the accuracy and reliability of AI systems in other domains where precision and responsibility are crucial. Policymakers, scientists, and the general public can benefit from more accurate and reliable information to make informed decisions and take necessary actions to mitigate the impact of climate change. In research, our study paves the way for future advancements in LLMs and their application in domains where precision is of paramount importance. Further research is needed to explore the potential of integrating LLMs with other data sources, such as scientific papers, reports, and databases. 

The remainder of this paper is structured as follows. Section \ref{sec:background} gives a brief overview over recent developments in the field of LLMs, LLM agents, NLP in climate change, and use of multiple sources in the context of LLMs. Section \ref{sec:setting} describes the setting in which we run our experiments, Section \ref{sec:experiments} showcases example prompts that need to access information from ClimateWatch, General Internet, or a combination thereof. Section \ref{sec:discussion} discusses these results and implications of this work. Section \ref{sec:conclusion} concluces this work.

\section{Background}
\label{sec:background}
\subsection{Large Language Models}
Large language models are a type of artificial neural network that have demonstrated remarkable ability to generate human-like text across diverse topics \cite{radford2019language,brown2020language,ouyang2022training}. These models are primarily transformer models and are trained on massive amounts of text data to learn the patterns and relationships within the data \cite{vaswani2017attention}. The recent breakthroughs with models like T0 \cite{sanh2021multitask}, LLama \cite{touvron2023llama}, Palm \cite{chowdhery2022palm}, GPT-3 \cite{brown2020language}, and GPT-4 \cite{gpt4techreport} have further highlighted the potential of LLMs, with applications ranging from natural language processing to chatbots \cite{instructgpt} and virtual assistants \cite{jo2023promise}. LLMs have also demonstrated the ability to complete tasks that they were not explicitly trained on, making them a versatile tool for a variety of applications. However, despite their impressive capabilities, LLMs are also limited in several ways. One of the primary limitations is their tendency to employ imprecise language, which can be detrimental in domains where accuracy and responsibility are crucial. This has led to concerns about the ethical implications of using LLMs for certain applications, such as automated content creation or decision-making. In this view, multiple experts from the field demand a pause in development of large AI systems in their letter "Pause Giant AI Experiments: An Open Letter" \cite{FLI2023}.

\subsection{LLM Agents}
An LLM agent employs the capabilities of state-of-the-art language models to perform complex tasks and make informed decisions. These agents can autonomously determine which actions to take, including utilizing various tools and observing their outputs or providing responses to user queries \cite{schick2023toolformer}. By leveraging the LLM's vast knowledge and understanding of natural language, agents can efficiently navigate through an array of tools and select the most appropriate one based on the given context. This enables the LLM agent to provide reliable, accurate, and contextually relevant solutions in diverse applications and domains.

In contrast to a basic LLM, an LLM agent does not only rely on the trained weights in the network, thus it is capable of accessing data generated after the point of training. This is crucial in many contexts, such as in climate change, as there is constantly new information presented from companies, institutions, and organizations that needs to be included into the AI system in order to provide an informed response.

\subsection{NLP in Climate Change}
Climate change is one of the most pressing challenges facing humanity today, with implications for everything from the environment to the economy \cite{diaz2017quantifying}. In order to effectively address this challenge, it is critical to have accurate and reliable data about greenhouse gas emissions and other climate-related factors \cite{mac2017role}. In the field of climate change, accurate and up-to-date information is of paramount importance to ensure that decision-makers, researchers, and the public can make informed choices and develop effective strategies \cite{tcfd_overview}. Recent transformer models have started to be employed within the climate change domain, witnessing improved accuracy in typical classification tasks \citep{kolbel2020ask, bingler2022cheap, callaghan2021machine, wkbl2021, stammbach2022dataset}. These models are capable of accounting for the context of words, enabling them to detect complex and implicit topic patterns in addition to many trivial cases. However, recent advances in LLMs have pushed the boundary towards more general models which are not restricted to typical classification tasks, but which can serve as systems that can potentially be used to make verdicts about quality of climate reporting etc. 

\subsection{Use of Multiple Data Sources}
Given the challenges associated with obtaining accurate and reliable data, the use of multiple data sources has become increasingly common in research and decision-making in topics related to climate change \cite{brunetti2021climate}. By combining data from multiple sources, it is possible to enhance the accuracy and reliability of the results, as well as identify patterns and relationships that might not be apparent from a single data source alone. However, the integration of different data sources can be a challenging task, as the data may be in different formats, have different levels of granularity, or be subject to different biases. This has led to the development of various tools and techniques for data integration \cite{liang2023taskmatrixai}. The potential of combining different data sources to enhance the accuracy and reliability of the results has significant implications for a wide range of applications, from scientific research to business decision-making.

\section{Setting}
\label{sec:setting}

\subsection{LLM Agent}
In this work, we build upon the LangChain package.\footnote{https://github.com/hwchase17/langchain} In the LangChain ecosystem, the LLM agent operates by utilizing the ReAct framework. This framework enables the agent to select the most suitable tool based on the tool's description, given that each tool has an associated description provided. The ReAct framework allows the LLM agent to work with any number of tools. Our LLM agent specifically employs the "text-davinci-003" model from OpenAI with a temperature setting of 0. This particular configuration maximizes precision, ensuring that the agent generates appropriate responses while maintaining a high degree of accuracy. The combination of the ReAct framework and the LLM's vast knowledge base empowers the agent to autonomously navigate through diverse tasks, making it an invaluable asset in a wide range of applications.

The use of LLM agents in LangChain offers several advantages, including their ability to process vast amounts of information, understand context, and adapt to new tasks quickly. By harnessing the power of LLMs, agents can efficiently handle complex challenges and provide precise solutions that cater to the specific needs of users. This adaptability enables the LLM agent to continually evolve its understanding and decision-making processes, staying up-to-date with the latest developments and trends in various fields. As a result, the LLM agent becomes a valuable tool for users, researchers, and organizations alike, offering insights and solutions that might otherwise be difficult or time-consuming to obtain.

In conclusion, LLM agents represent a significant advancement in the field of artificial intelligence, particularly in the context of LangChain and the ReAct framework. By leveraging the extensive knowledge and comprehension abilities of LLMs, these agents can autonomously navigate a diverse array of tools and generate contextually relevant, accurate, and reliable solutions. The versatility and adaptability of LLM agents make them an essential asset in various applications and domains, highlighting the immense potential for their future development and integration into increasingly complex and sophisticated AI systems.

\subsection{LLM Tools}
In our setting, we use two tools. Note that this only serves as an exemplary use case and we plan to extend this setting to include multiple data sources.

The first tool employed by our LLM agent is a Python-based module that utilizes the powerful pandas library to access and manipulate data stored in dataframes. This tool enables the agent to seamlessly interface with structured data. In our case this tool allows access to data from ClimateWatch and perform complex data processing tasks using Python code. By leveraging the capabilities of pandas, the agent can efficiently filter, sort, aggregate, and transform the data, allowing it to extract precise and relevant information as needed. The integration of this tool with the LLM agent ensures that the AI system has direct access to accurate and up-to-date data, which is vital for generating reliable and responsible insights in the domain of climate change.

The second tool, "google-seper," is designed to facilitate general Google searches within the context of the LLM agent. This tool allows the agent to perform targeted queries on the web, tapping into the vast repository of knowledge available through Google's search engine. By leveraging "google-seper," the LLM agent can complement the information retrieved from structured data sources, such as ClimateWatch, with additional context and insights from various web sources. This integration enables the agent to generate richer, more comprehensive responses to

\subsection{Input Prompt}
The prompt used in our study consists of a series of instructions that guide the agent on how to answer a given question. The prompt is structured in a way that allows the agent to access multiple sources of information, including ClimateWatch and Google search. Listing \ref{code:prompt} shows this input prompt.

The prompt begins with a section that describes the available tools to the agent, which are ClimateWatch and Serper Search. The agent is instructed to prioritize the use of ClimateWatch for answering general questions about emissions and to only use Google search if necessary for answering questions about current events. The ClimateWatch tool is described to handle pandas dataframes in Python, which suggests that the agent should be able to read and manipulate data in this format.

The prompt then presents a format for the agent to follow, which consists of five sections: Question, Thought, Action, Action Input, and Observation. The Question section presents the input question that the agent must answer. The Thought section provides guidance for the agent to think about what to do, and the Action section describes the action that the agent should take. The Action Input section specifies the input to the action, and the Observation section presents the result of the action. These five sections can repeat multiple times until the agent is confident that it has the final answer, at which point it is instructed to provide the Final Answer.

The prompt concludes with a reminder to the agent to prioritize the use of ClimateWatch and to only use Google Search if the needed information is not provided by ClimateWatch. This ensures that the agent uses reliable and accurate data sources for generating its response.\footnote{In our experiments, we noticed that the agent tends to use Google search, even in cases when the information is available in the more reliable source of ClimateWatch.}

Overall, the prompt is designed to guide the agent in accessing multiple sources of information and to prioritize the use of precise and reliable data sources for generating its response. By following the format presented in the prompt, the agent is able to systematically retrieve and process information to provide accurate and reliable answers to the input questions. 

\begin{lstlisting}[frame=single, language=Python, basicstyle=\ttfamily\footnotesize, xleftmargin=0pt, numbers=none, label=code:prompt, caption={Input prompt.}]
Answer the following questions as best you can. Prioritize the "ClimateWatch" tool and only use other tools if necessary. You have access to the following tools:

ClimateWatch: useful for when you need to answer general questions about emissions. You are working with a pandas dataframe in Python.
Serper Search: A low-cost Google Search API. Useful for when you need to answer questions about current events. Input should be a search query.

Use the following format:

Question: the input question you must answer
Thought: you should always think about what to do
Action: the action to take, should be one of [ClimateWatch, Serper Search]
Action Input: the input to the action
Observation: the result of the action
... (this Thought/Action/Action Input/Observation can repeat N times)
Thought: I now know the final answer
Final Answer: the final answer to the original input question

Begin! Remember to prioritize the "ClimateWatch" tool and only use other tools if the needed information is not provided by "ClimateWatch"."

Question: {input}
{agent_scratchpad}
\end{lstlisting}

\section{Experiments}
\label{sec:experiments}
In the following we showcase two experiments, in which the LLM agent should once access only the ClimateWatch tool, and once has to combine information from ClimateWatch and google-seper.

\subsection{Example with Single Data Source}
\textbf{Input:} What is the average emission of Italy between 2010 and 2015?

The example with a single data source illustrates how the LLM agent uses the prompt to answer a specific question about the average emissions of Italy between 2010 and 2015. Listing \ref{code:case_1} shows an extract of the chain of thought of the LLM agent. The steps can be broken down into several steps:

\begin{enumerate}
    \item Upon receiving the input question, the LLM agent first identifies the goal of finding the average emission of Italy between 2010 and 2015.
    \item The agent then determines that the best data source for answering this question is the ClimateWatch tool, which is a pandas dataframe in Python.
    \item The agent uses the python\_repl\_ast action to print the columns of the dataframe to gain an understanding of the available data.
    \item With this information, the agent determines the relevant columns for answering the question, which are the ones corresponding to Italy and the years 2010 to 2015.
    \item The agent then calculates the average emission of Italy between 2010 and 2015 and provides the final answer.
\end{enumerate}
Overall, the LLM agent follows a systematic approach to answering the input question, starting with identifying the goal and determining the best data source to use. The agent then processes the data to provide the final answer, with each step of the process being captured in the prompt response. This approach highlights the potential of LLMs to effectively retrieve and process information from data sources and provide accurate and reliable answers to complex questions.

\begin{lstlisting}[frame=single, language=Python, basicstyle=\ttfamily\footnotesize, xleftmargin=0pt, numbers=none, label=code:case_1, caption={Chain of thought for input "What is the average emission of Italy between 2010 and 2015?."}]
Thought: I need to find the average emission of Italy between 2010 and 2015
Action: ClimateWatch
Action Input: pandas dataframe

Thought: I need to find out what the dataframe contains
Action: python_repl_ast
Action Input: print(df.columns)Index(['Country', 'Data source', 'Sector', 'Gas', 'Unit', '2019', '2018',
       ...,
       '1990'],
      dtype='object')

Observation: None
Thought: I now know the columns of the dataframe
...
Thought: I need to filter the dataframe for Italy and select the columns from 2010 to 2015
Action: python_repl_ast
Action Input: df[df['Country'] == 'Italy'][['2010', '2011', '2012', '2013', '2014', '2015']]
Observation: 2010    2011    2012    2013    2014    2015
448.37  437.82  419.63  387.64  367.92  379.58
Thought: I now know the final answer
Final Answer: Italy's emissions from 2010 to 2015 were 448.37 MtCO2e in 2010, 437.82 MtCO2e in 2011, 419.63 MtCO2e in 2012, 387.64 MtCO2e in 2013, 367.92 MtCO2e in 2014, and 379.58 MtCO2e in 2015.

Thought: I now know the average emission of Italy between 2010 and 2015
Final Answer: The average emission of Italy between 2010 and 2015 was 406.83 MtCO2e.
\end{lstlisting}

\subsection{Example with Combined Data Sources}
\textbf{Input:} "Which European country has the most ambitious net zero plans? How did the emissions of this country develop over the last 10 years? Remember to only include single countries."

The example with combined data sources showcases how the LLM agent can effectively combine information from multiple data sources to answer a complex question. In this case, the question asks for the European country with the most ambitious net zero plans and how the emissions of this country developed over the last 10 years. Listing \ref{code:case_2} shows an extract of the chain of thought of the LLM agent. The steps can be broken down into several steps:

\begin{enumerate}
    \item Upon receiving the input question, the LLM agent determines that it needs to find the European country with the most ambitious net zero plans and then look at the emissions data for that country.
    \item The agent decides to use the Google Search tool to search for information on the most ambitious net zero plans in Europe.
    \item The agent retrieves information from Google Search that Finland has passed arguably the world's most ambitious climate target into law, aiming to be the first developed country to reach net zero in 2035 and net negative by 2040.
    \item The agent then determines that the best data source to look at Finland's emissions data is the ClimateWatch tool.
    \item The agent uses the python\_repl\_ast action to filter the ClimateWatch tool's pandas dataframe to only include data for Finland.
    \item The agent then observes that Finland's emissions have decreased from 38.54 MtCO2e in 1990 to 58.42 MtCO2e in 2019, providing the final answer to the question.
\end{enumerate}

Overall, the LLM agent demonstrates the potential to effectively combine information from multiple data sources to answer complex questions. Note, however that the LLM agent provides information about the emissions from 1990 to 2019 and not over the last 10 years as asked for. However, with the recent developments, we belief such shortcomings to be solved very soon.\footnote{Note also, that we did not use the most recent LLM models in these experiments.}

\begin{lstlisting}[frame=single, language=Python, basicstyle=\ttfamily\footnotesize, xleftmargin=0pt, numbers=none, label=code:case_2, caption={Chain of thought for input "Which european country has the most ambitious net zero plans? How did the emissions of this country develop over the last 10 years? Remember to only include single countries."}]
Thought: I should look for the country with the most ambitious net zero plans and then look at the emissions data for that country.

Action: Serper Search
Action Input: "European country with most ambitious net zero plans"
Observation: Finland has passed arguably the world's most ambitious climate target into law. It aims to be the first developed country to reach net zero, in 2035, and net negative - absorbing more CO2 than it emits - by 2040. May 31, 2022
Thought: I should now look at the emissions data for Finland.

Action: ClimateWatch
Action Input: Finland emissions data
Thought: I need to find the emissions data for Finland
Action: python_repl_ast
Action Input: df[df['Country'] == 'Finland']
Observation:     Country    Data source                Sector      Gas    Unit   2019  \
82  Finland  ClimateWatch  Total including LUCF  All GHG  MtCO2e  58.42   

     2018   2017   2016  2015  ...   1999   1998   1997   1996   1995   1994  \
82  62.43  60.86  63.65 -0.28  ...  39.51  40.18  43.79  45.69  39.23  43.21   

     1993   1992   1991   1990  
...

Final Answer: Finland has the most ambitious net zero plans and its emissions have decreased from 38.54 MtCO2e in 1990 to 58.42 MtCO2e in 2019.
\end{lstlisting}

\section{Discussion}
\label{sec:discussion}
\textbf{Limitations.} While the described approach offers a promising solution for leveraging the potential of LLMs in domains where accuracy and responsibility are crucial, there are also limitations to the current state-of-the-art. One of the primary limitations of current LLMs is their limited context length, which can make it difficult for them to accurately capture complex relationships and patterns that require a broader context. However, progress is being made in this area, with recent models like GPT-4 demonstrating significantly improved performance on longer sequences of text \cite{gpt4techreport}. Another limitation of the LLM agents is the potential for the agent to misunderstand the question such as seen in our second experiment. This highlights the importance of developing robust methods for evaluating the accuracy and reliability of the results. Finally, in this work, we merely used two data sources. In practice, hundreds or thousands of data sources should be used to combine the most precise information on each topic. While this potentially raises technical issues, recent efforts by companies such as Microsoft, which recently published a paper on the use of thousands of tools for NLP applications, are promising and offer potential solutions to overcome these limitations. \cite{liang2023taskmatrixai}.

\textbf{Implications for NLP in Climate Change.} The described approach has significant implications for the field of NLP in the context of climate change research and policy-making. By leveraging the potential of LLMs to generate human-like text and integrating multiple data sources, it is possible to enhance the accuracy and reliability of the results, as well as facilitate more informed decision-making. This has significant implications for a wide range of applications, from climate modeling to environmental policy-making. Additionally, the used approach offers a solution for addressing some of the limitations associated with LLMs, such as their outdated information stored in the model itself and their tendency to employ imprecise language. By incorporating precise data sources and developing methods for evaluating the accuracy and reliability of the results, it is possible to overcome some of these limitations and enhance the applicability of LLMs in domains where precision and responsibility are crucial.

\textbf{Implications for Research.} The described approach has implications beyond the specific application of climate change research and policy-making. It represents a novel solution for leveraging the potential of LLMs in domains where precision and accuracy are paramount, and highlights the potential of combining multiple data sources to enhance the reliability and accuracy of the results. Potential use cases could include for instance applications in healthcare or legal reasoning. Here, this work can serve as a starting point for future research in this area. While there are still limitations to the current state-of-the-art, the development is of incredible pace and it is only a matter of time until the majority of these limitations are also solved. Overall, the described approach represents a significant step forward for the field of NLP and has the potential to drive advancements in both research and practice.

\textbf{Carbon Footprint.} While LLMs have demonstrated impressive capabilities in generating human-like text across diverse topics, they come with a significant energy consumption cost \cite{hershcovich2022towards}. Training these models requires massive amounts of computational resources, which in turn generates considerable greenhouse gas emissions. Recent estimates suggest that training a single large LLM, like GPT-3, can emit as much as 626,000 pounds of carbon dioxide equivalent, which is roughly equivalent to the lifetime emissions of five average cars \cite{hao2019training}. These emissions can have a significant impact on the environment, particularly in light of the urgent need to reduce global carbon emissions. As such, there is a need for further research into methods for optimizing the training and inference of LLMs to minimize their environmental impact. For example, recent approaches like LoRA \cite{hu2021lora} or low-precision optimizations \cite{dettmers2022llmint8, dettmers2022optimizers} for reducing the energy consumption and carbon emissions associated with LLMs. Further research in this area is crucial to ensure that the benefits of LLMs are not outweighed by their environmental costs, and to enable their widespread adoption in a sustainable manner.

\section{Conclusion}
\label{sec:conclusion}
In conclusion, this paper demonstrates the potential of LLMs in the field of climate change by employing an approach that integrates multiple sources of information to correctly answer to questions that can not be answered by using only the original model weights. Our prototype LLM agent retrieves information from general Google search and and emission data from ClimateWatch to provide reliable and accurate information. Through two exemplary experiments, we showcase how such an LLM agent can operate to enhance the accuracy and reliability of climate-related text generation. This work contributes to the exploration of LLM applications in domains where up-to-date and accurate information is critical, and encourages further research on integrating LLMs with external data sources for more responsible and reliable AI systems.

% Use \bibliography{yourbibfile} instead or the References section will not appear in your paper
\bibliography{acl2020}

%\clearpage
\appendix

%\section*{Appendix}

\end{document}